# Impact of automation during innovative remanufacturing processes in circular economy: a state of the art

Perla Nohra
Grenoble INP, CNRS University of
Grenoble Alpes, G-SCOP,
46 avenue Félix Viallet
Grenoble, 38000, France
perla.nohra@grenoble-inp.org

Helmi Ben Rejeb
Grenoble INP, CNRS University of
Grenoble Alpes, G-SCOP,
46 avenue Félix Viallet
Grenoble, 38000, France
helmi.ben-rejeb@grenoble-inp.fr

Swaminath Venkateswaran
Léonard de Vinci Pôle
Universitaire,Research Center,92916
Paris, La Défense, France
swaminath.venkateswaran@devinci.fr

*Abstract*— **With the increasing demand of raw materials nowadays, and the decrease in supplies, the industrial sector is suffering. The environment and the society are also indirectly affected. The goal to reach a sustainable development imposes several studies on the economic, environmental and community level. The aim of this paper is to provide an overview of the existing body of literature on automated remanufacturing, and its potential impacts on the three pillars of sustainability. A particular interest is given to the growing use of cobots promoted by the principle of industry 4.0. The investigation that covers each part of the remanufacturing process will help in formalizing an approach about the automation of such processes. It highlights the challenges found and aims to improve the remanufacturing sector towards a more sustainable industry.**

## I. INTRODUCTION

The global population continues to grow rapidly, with greater demand for raw materials. This has led to enormous energy consumption [1]. The supplies are decreasing, imposing an alert in the industrial sector. In this context, conserving resources and reducing waste stream is urgent to reach environmental protection, society equity, and economic viability [2]. Hence, the increasing environmental burden has been prompting governments worldwide to seek for more sustainable economic models. Increased attention is being given to circular economy (CE) based on resource conservation and material flow valorization through the 3R's principle, labeled as Reducing, Reusing, and Recycling [3]. The depletion of natural resources is increasing, leading to degradation. The time needed to replenish these resources will enlarge the global eco-footprint [1], [4]. We are facing this shortage due to overpopulation, pollution, overconsumption, industrial and technological development. As more countries achieve big technological advancements, the modern world continues to become increasingly industrialized. The technological breakthroughs accelerate, so does the growth of industries that demand raw materials for development and production, hastening the shortage of resources [4]. Therefore, the development of new strategies to reduce, reuse, and recycle materials is necessary.

Thus, to maintain the value of resources, and increase their efficiency, the circular economy (CE) paradigm is crucial. Based on resource conservation and material flow valorization through reducing, reusing, and recycling, the CE concept aims to decrease the dependence on major resources. Nonetheless, the path towards implementing CE is still long [5]. The 3Rs principle contribute to a closed loop where the benefits are manifold. Reducing contributes to decreasing waste generation. Reusing helps in less material extraction. Recycling transforms the waste itself into a new resource and usable material. So, it is a better alternative to the adopted system of economy based on take–make–dispose [6]. This reveals the role of the remanufacturing process as a key enabler for more sustainable manufacturing [7]. It aims to closing natural resources flow while optimizing the other values throughout the lifecycle of materials [8]. Fig.1 shows the loop of CE, where instead of disposing away a product as it approaches the end of its useful life. The materials are preserved so they can be used again and again, providing additional value [9]. Several challenges that may prevent or slow down the implementation of CE have been recognized [3]. The lack of reliable data and information, shortage of advanced technologies, weak or absent economic incentives, and even customer's satisfaction are challenging. Also, poor enforcement of legislations, leadership and management, lack of public awareness, and lack of a standard system for assessing CE's performance are considered from the main limitations [3].

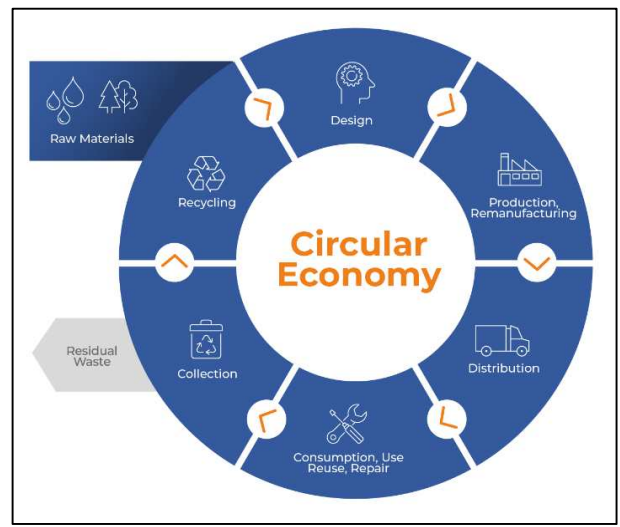

Fig. 1. Circular economy illustration [9]

However, the CE perspective is intended to regenerate sustainability capital. It aims to achieve greater results for future generations by cascading components for the longest lifetime [10]. This mode of economic development fits perfectly in the industrial field and could be strongly connected to the concept of remanufacturing. Knowing that many experts are now looking into closed loop supply chain solutions like remanufacturing that can help achieve the CE goal [11]. The resulting benefits here are not limited in the industrial frame but contribute also to the overall sustainability. All the pillars are affected, since they are strongly linked together, and Fig.2 shows this connection.

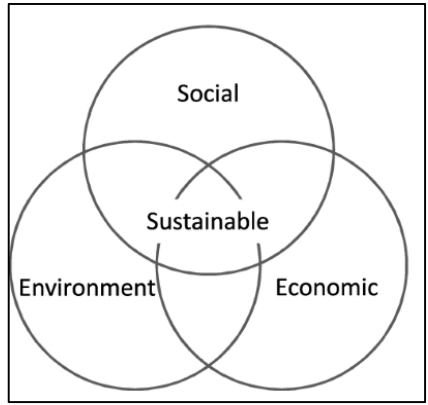

Fig. 2. Sustainability dimensions [12]

Therefore, closing the loop, fits perfectly with the needs of the industrial field nowadays. This imposes the investigation of the remanufacturing process in the industry. Several studies highlighted the benefits of remanufacturing and its potential impacts on sustainability expressed as the combination of three pillars: social, economic, and environment [13], [14]. Positive impact was determined on the social level in terms of employment rate, creation of new jobs, gender involvement, and consumer satisfaction rate [13]. On the economic level, remanufacturing allows cost efficiency, material recovery, and productivity [15]. On the environmental level, it allows the reduction of energy consumption and waste disposal [16].

Moreover, the automation of remanufacturing is a new challenging scenario.[17]. Remanufacturing is often an activity which includes various manual operations, that are often repetitive and exhausting for the operator. These operations are prone to human error and might have inconsistency in quality [17]. Thus, a fully automated remanufacturing process has not yet been achieved. The automation of operations plays an increasingly important role in remanufacturing today [18]. The major player in this automation is the cobot, due to its collaborative features [19]. Nevertheless, it has some limitations in terms of its ability to perform all the operations of the remanufacturing process [17].

In this paper, we will shed the light on the relationship between remanufacturing, sustainability, and automation. The interdependency between these terms will be shown. The investigation covers the criteria that are being considered in literature to determine the operations of the remanufacturing process and the challenges that may encounter its automation.

The objective of the presented research is to give an extensive view from the literature about the use of cobots in automating remanufacturing and its contribution to CE. We have followed a systematic literature review based on the PRISMA methodology to determine the most relevant papers that cover our research area, presented in section 2. Section 3 presents the outcomes of this literature review. It allows the identification of the main concepts investigated, describes the stages and the benefits of the remanufacturing process, and investigates the use of cobots. The challenges that might prevent a successful automated operation are also highlighted. In section 4, the results are discussed, highlighting gaps in the literature. Finally, Section 5 is the conclusion and gives future research perspectives.

## II. METHOD

To meet the aim of the paper, a systematic literature review was performed to gather all the available data covering the remanufacturing topic, its automation, and its impacts on sustainability. It aims to improve the research performance and the quality of review processes [20]. The PRISMA methodology was used, to minimize research bias and maximize reliability and replicability [21]. It allows readers to assess the strengths and weaknesses and demonstrates the quality of review. Thus, it permits the replication of review methods through its 4 phases flow diagram [22].

The literature search used words including *Remanufacturing, Sustainability, Circular economy, Cobots, automation tools*. Web of science and Google Scholar databases were used. These keywords were refined, to get broaden and useful results. A reference management software was used to organize them and delete duplicates.

The 4 phases of PRISMA methodology consist of: identification, screening, eligibility check, and finally included articles. The first step covers all the records in the research area, from all the databases. All the papers found in this step are gathered in a library. The screening is a step of filtering, where the duplicates are removed. The eligibility is the examination phase. The records here are all exposed to categories selection. The aim is to remove the records that do not have a high match with the topic. The selection criteria are important to narrow the research coverage, and to get to the core. For example, in this case, the records covering the civil engineering field were excluded. Many other fields are out of scope, so they are not eligible. Finally, the records included are selected based on their contribution to the topic. Thus, to reveal the connection between the references included, and to highlight the focus on these topics, a map generated by VOS viewer is shown in Fig. 4. VOS viewer is a software that creates maps based on bibliographical data [23]. The map creates a network of bibliographic sources based on the keywords or tags co-occurrence (i.e., the number of keywords present simultaneously in two different bibliographic sources). The input of the map is exported from the bibliography database file in RIS format. After the creation of the map, it is possible to explore it with several features for visualization. In our case, the network visualization is selected. Keywords are labeled and, by default, circle are used to represent them in the network. The weight of an item determines the size of the label and the circle around it. The larger the label and the circle of an object, the greater the weight of the item. Links are represented by lines.

After applying the PRISMA method and the representation of co-occurrence map with VOS viewer, the results are discussed to determine main patterns in the literature regarding remanufacturing and automation.

## III. RESULTS

Fig. 3 shows the identification, screening, eligibility, and the included articles in the literature, highlighting the research interest in this topic. This reveals the great background available and the successful matching between our keywords.

This initial systematic review yielded 5 201 papers. The number of records, decreased from 5 201 to 4 324 after duplication The recorded number decreased from to 685, after removing the records that does not match the scope. The exclusion phase was done through removing the records related to Civil, and metallurgical engineering, architecture, thermodynamics, computer sciences, and many more sectors. After achieving a great number of eligible records, filtering again is necessary. The filtering was based on a quick analysis concerning the cobots. The focus on using collaborative robots, which are the cobots, started mainly with the industry 4.0. This means that the research area must be limited to the year 2010. After screening again and narrowing the parameters of selection, 56 records were included in the review.

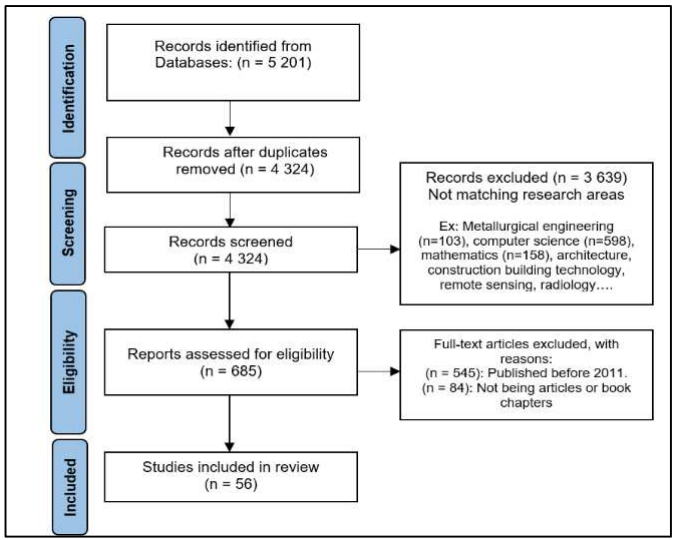

Fig. 3. PRISMA flow diagram

### A. Findings from Systematic Literature

Among the 56 references included in our research, the most common keywords are remanufacturing, sustainability, cobots, and circular economy. The records that investigated sustainability, covered also all the dimensions as pillars. However, the map of Fig. 4 yielded by VOS viewer shows the interest given to these topics.

According to this figure, there is a great connection between automation and remanufacturing. So, with the remarkable increase of technology and artificial intelligence, the automation is found to satisfy essential needs in the industry and ease mass production. Hence, we cannot deny the research interest about the uncertainty in the remanufacturing process, the strategies, and even the process assessment. The co-occurrence of these terms in many references shows that the community of researchers are interested in investigating this topic with all its pros and cons. So, balancing between sustainability and automated remanufactured can bring profitability for adopting CE which directly affects the sustainability development.

Remanufacturing is the process of returning a product to a serviceable condition. It deals with the disassembly, cleaning, inspection, restoration, and reassembly. [24]. The resultant product is guaranteed to be equivalent to a newly manufactured one with the same quality and performance [7].

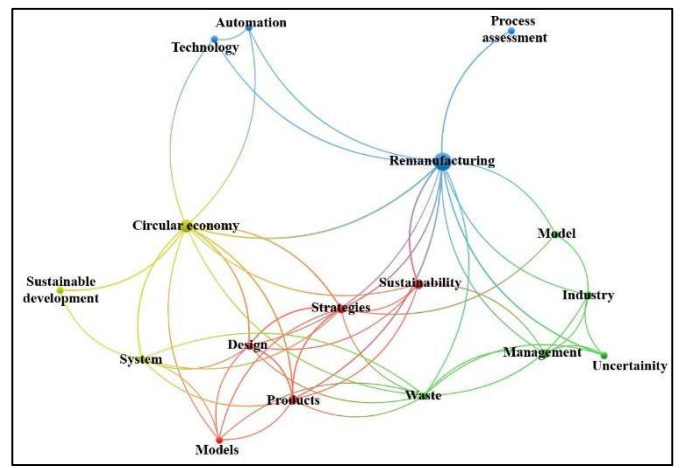

Fig. 4. VOS viewer map of keywords co-occurrence

Engineering parts exposed to harsh environmental conditions, wear, cracks, deformation, and long hours of operations can be remanufactured. These defects can affect the overall performance, so, it is necessary to extend their life cycle. Remanufacturing is profitable because it is done on high-value engineering parts such as impeller, shafts, turbine blades, and can cover different fields [25]. However, to reach the desired outcomes of the remanufacturing, the challenges faced are numerous: high labor cost, inconsistency in remanufactured parts quality due to frequent abnormalities, and complexity in the shape of components [26]. Unfortunately, there are not any defined tools or systems available to analyze and compare remanufacturing processes, since it is not a globalized approach yet. Thus, to increase the quality and the expected outcomes, some steps are vital to follow, where a model for assessment was validated [27]. Somehow, to overcome the technical problems and barriers in such process, many concepts were proposed such "Design for remanufacturing" [7]. This consists of analyzing the probable faced barriers in the early design phase, where some parameters could be changed or improved.

However, the CE concept is not formally and completely introduced in the academic base for engineers, so future engineers are not getting prepared to incorporate it in their design at first, which is considered as a later-on challenge. This requires governments to act and legislate.

Thus, a study comparing the design of a product relying on linear economy, and the redesign of the same product using the circular economy model, while using the sustainability indicators was done to showcase the benefits of the CE model [28]. So, by means of direct relation between remanufacturing and sustainability, the CE model is highlighted and incorporated in the Life Cycle Analysis (LCA), which is another point of interest, that shows value for remanufacturing. The standards and guidelines of LCA need to be followed [29] since it covers many decisions and applications and take into consideration all the constraints and burdens related to the product.

All these concepts, combined, find fit in the strategic remanufacturing sectors, such as aeronautics, automotive, electronics, consumer goods, and mechatronics. And, to reach a successful process, enhancing the features of remanufacturing systems through higher levels of automation, flexibility and adaptability to materials is a concern [30]. It is a "win-win" concept that faces many challenges on the automation level. The remanufacturing process is three to five times more labor intensive than the manufacturing process of a product. The disassembly activities, cleaning, inspection, and sorting are not present in manufacturing [26]. So, this implies the need for higher degrees of automation.

Each phase of the remanufacturing process should be investigated and discussed to understand its range of adaptability to automation. [31]. Thus, the robots are generally major players in automating any operation, but they don't fit within the steps of the remanufacturing process. Up till now, whatever was the degree of automation, a successful remanufacturing process, cannot be done without the human interference. So, the robots here are not encouraged, whereas cobots are.

Cobots positively influence the success of the remanufacturing process because they are way precise, accurate, and consistent. Time, power, and money wise, the cobots are taking a greater place in the industry nowadays.

As definition, cobots are collaborative robots with less safety restrictions than industrial robots. They can perform various tasks perfectly. They are inherently passive robots intended for direct collaborative work with a human operator. The controller guides motion by tuning the cobots set of continuously variable transmissions [32]. That's why, they are widely used in the remanufacturing industry since they minimize the risk of accidents. Cobots allow the employees to have more time for other applications, so, the productivity increases.

Thus, the adoption of cobots for automating remanufacturing is advantageous. The tasks are constant, repetitive, frequent, rerunnable, and parametrized. And most importantly, they are granular, so the lowest of level of automation can be set up for manual intervention. The efficiency of cobots is undeniable, but many limitations are present: cost, time, safety, and applicable operations. Whereas the aim in each operation is to reach high flexibility, intellectualization, precision, and integration.

Therefore, cobots can fit perfectly to perform most of the steps of the remanufacturing but they will not fit with all types of products. The complex subassemblies can limit the process. There have been many investigations in this field and several disassembly sequence planning methods have been developed, to make this step faster and easier than it's used to be [33]. Moreover, hundreds of investigations and studies were done to improve the disassembly phase, but many challenges are encountering the realization of the process. This emphasizes more technologies and more features to be incorporated within the cobots functions and applications.

Industry 4.0 is increasingly being promoted as the key to improving productivity, promoting economic growth and ensuring the sustainability of companies [34]. The automation itself is not sustainable. It needs energy, and raw materials to be developed [35], but it contributes to achieving sustainable goals. The relationship between the automation and the industry is inter-related due to the need of mass production to satisfy the economic growth.

The key is to balance between the cost of the automation process and its outcomes. The fast-growing technologies and the artificial intelligence, AI, are interwinding the relation between human and robots, which is becoming more collaborative, not competitive [36]. This explains the on-going direction towards Industry 5.0.

*B. Key concepts*

- *Background of the increasing momentum of remanufacturing*

Compared with the original manufacturing adopted by industries, the remanufacturing can save 50% of the cost, 60% of the energy, 70% of the material, and 80% of the air pollutant emissions [37]. For this reason, the companies are now oriented towards adopting remanufacturing and developing their strategies in terms of sustainability development. Such comparisons give remanufacturing a greater place in the circular economy model, where the main three ideologies are the 3Rs by means of raw materials, manufacturing, and final products. Hence, industry, economics, society, and environment are closely inter-related. Therefore, the remanufacturing industry is shown to be a typical high efficiency economy, which imposes the translation to a complete Circular Economy concept [38]. The global remanufacturing industry output in 2013 has exceeded \$130 billion, which has included several profiles such as construction machinery, medical equipment, office supplies, and automobiles. And, according to a recent study, switching to a CE model might provide the European economy yearly an amount of 0.9 trillion Euros by 2030 [39]. So, the economic benefit and the revenues of the remanufacturing industry are huge. Moreover, the future promises more since the sustainability development goals are becoming the new standards for any company.

Based on what was mentioned before, the adoption of remanufacturing systems is emerging, with more interest to reach better quality and flexibility. The innovation of automating this process through the cobots, and increasing its adaptability, and its controllability, contributes to the CE paradigm's success [39].

- *The urge to adopt remanufacturing*

The circular economy has undeniable long-term sustainability promise [40]. Its realization and adoption in the industry is very crucial. It can open new business prospects while lowering material prices and reducing price volatility [41]. The CE concept encourages restorative and regenerative design, knowing that the linear economy concept is not sustainable anymore. The problem stands for some elements that could be depleted within the next 5–50 years, like gold, silver, indium, iridium, tungsten, and many others that are vital for industry. So, closing the loop, reducing this shortage, and increasing the reuse of materials can maintain a balanced society for the future generations. This could be done through remanufacturing. CE practices have potential to bring 80%–90% savings in raw materials and energy consumption, compared to the traditional linear model [30]. The general component price will be reduced of approximately 25%–30% which can boost competitiveness in emerging markets[39]. The availability of high-quality products, with a sustainable profile, and an affordable price can push the industries to adopt the remanufacturing process.

*C. Definition, benefits, and steps of remanufacturing*

The remanufacturing concept consists of the repair and restoration of a product that has reached his end-life cycle. It aims to transform it into a brand-new product with affordable price. It addresses an added-value strategy to the product [42].

The outcome is perceived to be of high quality and can be evaluated through the following factors: features, lifespan, serviceability and performance [43]. The operations of this process are consecutively six: disassembly, cleaning, inspection, repair/restoration, reassembly, and testing. This cycle contributes in closing the loop of raw materials consumption and creates an efficient recovery strategy for the product [42].

Any used product, received for remanufacturing will follow the six steps process, that will be briefly discussed with the possible challenges that may be encountered.

1- Disassembly: during this step, the product will be disassembled, as an introductory step to the process. It involves general purpose tools, such as a power drill, and might requires some automation, depending on its geometry and its features. This step requires time and effort, especially in case of mass production in large industries, so its automation could be done using a cobot. The human intervention here will be limited to the selection of the adequate gripper and the control.

2- Cleaning: after disassembling all the components of the item, cleaning them is a major concern. In terms of quality, and even precision, the surface of any component must be completely cleaned. Removing rust and dust residues leads to a better surface finishing, so this component will be ready to get repaired or reprocessed. The automation of this step depends on the requirements. It could be done through water jet, blowing, or placing into a dissolvent.

3- Inspection: it is usually done visually, to check the degradation or the failure of any component. It summarizes the re-manufacturability status of the product and orients the following steps of the process. In this phase, the selection of the components that can be repaired, or the ones that are going to be replaced take place [31].

4- Restoration: in this step, each component will be reprocessed to reach its original condition. This operation is the longest, because it depends on the shape of each component and its complexity. The sub-components of the remanufactured product might need repair or replacement. Some components might need cutting, welding, or trimming operations, so it is important to set them up to their original condition [31]. And here the use of cobots plays a major role to increase the performance of these tasks.

5- Reassembly: after finalizing the restoration step, reassembling all the components together to get the final product. This is very similar to the disassembly operation.

6- Testing: usually done by labors, to test the quality and the performance of the outcome. The uncertainty in quality requires much attention to make sure that the standards set are reached.[44].

So, all these steps could be done effectively using the cobots. Reduced material waste, time and effort are achieved through automation even if many barriers are found. Closing the loop through remanufacturing is improved using cobots. And the automated remanufacturing will have potential impacts on sustainability.

- *Cobots use in remanufacturing*

According to previous studies [45],[46],[47], there are potential areas where automation can be adopted in the remanufacturing industry. Disassembly, cleaning, and reassembly were judged to be the most promising areas for automation [46]. In this case, the automation of these operations contributes in controlling the efficiency and flexibility that are caused by a variety of product kinds [48]. So, the advantages of the automation are conclusive and undeniable in the remanufacturing process. But, many challenges and several technical and management difficulties have been identified in the literature [49]:

- Difficulties in disassembly and reassembly.

- Uncertainty in processing time.

- Uncertainty about the quality of the outcome.

Hence, the typical disassembly procedure is completed by either low-efficiency manual disassembly or low-flexibility robotic disassembly, and it is the same for the reassembly phase [50]. For this reason, cobots are selected to achieve a great integration between the manual and automatic operations. To be able to conduct a successful remanufacturing process, the products must achieve some eligibility criteria, by means of features and characteristics [44]. The main criteria that have been considered in the literature are:

- The product consists of interchangeable parts.

- The added value due to remanufacturing is high.

- Easy to be processed, by means of geometry.

- Relative and affordable cost.

However, a study focused on the characteristics of cobots that make them suitable for remanufacturing. It covers their ability to achieve heavy duty tasks and mass production, or the possibility to pick multiple parts in one lift. Also, their ability to automate practically any manual process, particularly those involving tiny quantities or frequent changeovers [51].

## IV. DISCUSSION

### A. Research outcomes

Based on what was found, a solid background and interest regarding the history of the remanufacturing is present. This summed up the relation of remanufacturing and automation and highlight the potential positive effects on sustainability pillars. The importance of automating such process was revealed through the advanced options provided by collaborative cobots, and even its contribution to reach better outcomes on the social, economic, and environmental level. Each step of the remanufacturing process has its own constraints to consider. We cannot reach a fully automated operation if the remanufactured part has a complex shape. This is very common for the disassembly and reassembly operations. Consequently, the cleaning stage might also need human intervention, in case the part needs a specific surface finishing. So, it is all about investigating each phase alone to study its flexibility when advanced automation using cobots is adopted.

### B. Research gap

Based on this literature review, a new gap is found, which is about the effect of automation on remanufacturing; what are the characteristics of the products that can be remanufactured using the cobots? There are not enough studies and articles investigating this topic. The automation might have several advantages but not in all fields. It is accompanied with great challenges and constraints. In the short to medium term, disassembly cannot be carried out entirely by cobots. Manual works are still required to perform difficult tasks that involves high degrees of uncertainty [52]. So, a future study can analyze each phase of the process, to investigate the operations that still need human intervention. And future research can be directed to measure the potential impacts of automated remanufacturing on all the sustainability levels, through conducting environmental assessment, ergonomic studies, and economic investigations.

Moreover, the quality of the remanufactured product is another constraint to consider. It is becoming a bottleneck restricting the development of the remanufacturing industry. The parameters that must be taken into consideration to satisfy customers are personalization, diversification, and environmental protection [53]. If reached, at the same time, the remanufactured product, will have some uncertainty, but in return, it has a high repair rate. So, the uncertainties of the remanufacturing process should be understood, analyzed, and decoupled from a systematic perspective. The operations that can be performed in this frame should be revealed, to achieve a controllable, and visual global approach. It can be foreseen that the continuous progress of the theory, method, and technology of remanufacturing assembly will play an increasingly important role in the scale and industrialization of remanufacturing industry [54].

A rich theoretical literature can be applied to the question of how the remanufacturing is influencing the economic, environmental, and social aspects of sustainability, and how the CE will contribute to reduce our ecological footprint. Thus, when adopting the remanufacturing process, with all the facilities found nowadays, will the industry 4.0 be recording more evaluation in terms of sustainability, and would a product design be done based on this ideology?

## V. Conclusion

Our research identified relevant papers and resources to broaden the knowledge about the remanufacturing concept, but our aim is to globalize the use of cobots in remanufacturing. Based on what was found, a high interest to reach more sustainability is found, and the concepts to set strategies have been already launched. The pillars of sustainability are all affected by adopting remanufacturing. But the key turning point now is related to how remanufacturing is affected by automation. All these terms are closely inter-winded, so investigating and assessing their improvements is relevant. The automation is our main concern. So, investing in advanced technologies to improve sustainability is crucial. Covering the aspects of economy, society, and environment, will help us to formalize a general approach concerning remanufacturing within the CE paradigm.

Future research perspective could be oriented towards experimental testing to study which of the operations can be done automatically using the cobots and which cannot. The testing will showcase the operations that needs mixed intervention also. It will assess the impacts of remanufacturing process through life cycle analysis (LCA), CE paradigm, and social equity. The LCA will cover all the stages of the product, starting from the extraction of raw materials to manufacturing, distribution, repairing, until it reaches the stage of recycling and its final disposal. The social equity study will cover ergonomic studies. The economic impact will be assessed through studying the contribution of the process to the CE.

The reached sustainability must be defined within a globalized approach to measure it. So, for future research, we aim to formalize a general approach that can be adopted in all the industries interested in minimizing the footprint through remanufacturing. This approach could present a new perspective regarding the use of cobots in automating remanufacturing. It could give general direction about automating each phase of the process, and needs be implemented by any remanufacturing employee through a clear sequence of commands and actions. It should highlight the possible barriers that may encounter the cobot in performing the six steps of the remanufacturing process.